\PassOptionsToPackage{round,authoryear,sort&compress,super}{natbib} 
\documentclass[preprint,12pt]{elsarticle}
\pdfoutput=1

\usepackage{amssymb}
\usepackage[T1]{fontenc}
\usepackage{lmodern}

\usepackage[utf8]{inputenc} 
\usepackage[T1]{fontenc}    
\usepackage{hyperref}       
\usepackage{url}            
\usepackage{booktabs}       
\usepackage{amsfonts}       
\usepackage{nicefrac}       
\usepackage{microtype}      
\usepackage{ctable}
\usepackage{graphicx}
\usepackage{subfig} 
\usepackage{tabularx}

\usepackage{multirow}

\journal{Medical Image Analysis}
\begin{document}
\begin{frontmatter}



\title{Adversarial training with cycle consistency for unsupervised super-resolution in endomicroscopy}

%

\author[$^1$]{Daniele Rav\`{\i}}
\author[$^2$]{Agnieszka Barbara Szczotka}
\author[$^3$]{Stephen P Pereira}
\author[$^4$]{Tom Vercauteren}

\address[$^1$]{Centre for Medical Image Computing, University College London}
\address[$^2$]{Wellcome/EPSRC Centre for Interventional and Surgical Sciences, University College London}
\address[$^3$]{Institute for Liver and Digestive Health, University College London}
\address[$^4$]{School of Biomedical Engineering \& Imaging Sciences, King's College London}

\begin{abstract}
In recent years, endomicroscopy has become increasingly used for diagnostic purposes and interventional guidance. It can provide intraoperative aids for real-time tissue characterization and can help to perform visual investigations aimed for example to discover epithelial cancers. 
\textcolor{black}{Due to physical constraints on the acquisition process, endomicroscopy images, still today have a low number of informative pixels which hampers their quality.} Post-processing techniques, such as Super-Resolution (SR), are a potential solution to increase the quality of these images. SR techniques are often supervised, requiring aligned pairs of low-resolution (LR) and high-resolution (HR) images patches to train a model. However, in our domain, the lack of HR images hinders the collection of such pairs and makes supervised training unsuitable. For this reason, we propose an unsupervised SR framework based on an adversarial deep neural network with a physically-inspired cycle consistency, designed to impose some acquisition properties on the super-resolved images. Our framework can exploit HR images, regardless of the domain where they are coming from, to transfer the quality of the HR images to the initial LR images. This property can be particularly useful in all situations where pairs of LR/HR are not available during the training. Our quantitative analysis, validated using a database of 238 endomicroscopy video sequences from 143 patients, shows the ability of the pipeline to produce convincing super-resolved images. A Mean Opinion Score (MOS) study also confirms this quantitative image quality assessment.
\end{abstract}


\end{frontmatter}



\section{Introduction}
According to a recent report by the World Health Organization, cancer is the second leading cause of death after cardiovascular disease and was responsible for 8.8 million deaths in 2015. Early detection, such as the ability to detect precancerous lesions, plays an important role in reducing cancer incidence and related mortality~\citep{torre2016global}. Optical endomicroscopy, based for example on confocal microscopy, optical coherence tomography or spectroscopy, has the ability to perform optical biopsies and identify early pathology in tissues or organs including the colon, oesophagus, pancreas, brain, liver and cervix~\citep{ravi2017manifold,nguyen2015current}. Although, in the last years, progress has been made to build reliable optical endomicroscopy devices~\citep{neumann2010confocal}, the need to operate at micron scale through the use of endoscopes, fibre bundles, laparoscopes, and needles, limits the final resolution of the images. Further hardware improvements are difficult to achieve and one possibility to improve the image quality is to post-process the images using SR techniques.

Recent methods for SR are based on training example-based models that learn how to improve image resolution by exploiting a database of aligned pairs of LR and HR images~\citep{ledig2016photo,ravi}. Nonetheless, due to the lack of HR endomicroscopy images, these pairs are not typically available in this domain. An option is to generate these pairs synthetically, but achieving this in a sufficiently realistic manner is only feasible when the acquisition process is extremely well defined. In most of the cases, the acquisition process is only known approximately and supervised methods may thus not be applicable.

\begin{figure*}[t!]
  \begin{center}
     \includegraphics[width=0.90\textwidth,clip]{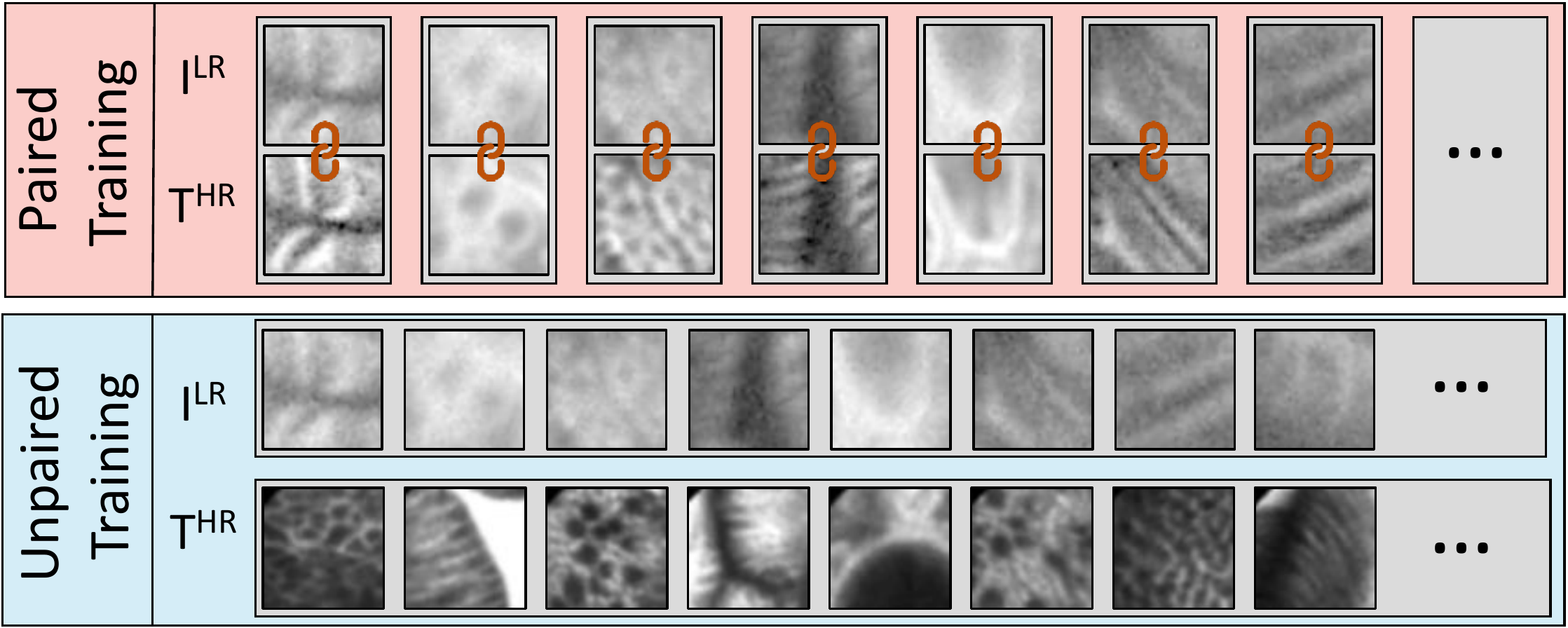}
    \caption{Example of aligned-paired and unpaired patches used for training super-resolution neural networks.}
    \label{fig:PairNoPair}      
\end{center}
\end{figure*}

For this reason, we designed a deep learning architecture trained in an unsupervised manner where the aforementioned one-to-one alignment between LR and HR is not required anymore. We formalize our framework so that LR images from an initial input domain $I^{LR}$ could be transformed into images of any target domain $T^{HR}$. The target domain can be the same or different from the initial one.
An example of the difference between initial and target domain with paired and unpaired patches is shown in Fig.~\ref{fig:PairNoPair}. 
To train the model using unpaired patches and avoid that the network learns to produce HR images with no direct relationship with the input images, a cycle consistency block is included in our architecture. This block imposes some physical acquisition properties so that the obtained HR images faithfully represent the initial LR images.

More specifically, during the training procedure, we make use of an adversarial network, a class of artificial intelligence algorithms which train two separate models that challenge each other in a zero-sum game. The first model is a $SR$ network that learns how to improve the resolution of the images, and the second is a discriminative network $DS$ that, looking at the target domain, tries to distinguish images generated by the $SR$ network from the real $T^{HR}$ images. The aim of $SR$ is to learn how to fool the $DS$ network and this leads to a generation of super-resolved images. 
\par
Adversarial training can learn how to produce outputs with the same distribution as the target domain. However, the target domain distribution could be sampled by simply mapping the input images to any random permutation of images in the target domain. Therefore, in this context and without specific constraints, an adversarial loss, alone, cannot guarantee that the learned function maps an input to a desired corresponding super-resolved image. Thus, following the idea proposed by~\citep{zhu2017unpaired} in the context of style transfer, we add in the adversarial training a further cycle block that imposes a consistency between the HR images and the initial LR images. In contrast to the work in~\citep{zhu2017unpaired} where the reverse mapping between the target and the source domain is also learned, in this work, this consistency is obtained by constraining the super-resolved $T^{HR}$ image to have similar physical acquisition properties to the initial $I^{LR}$ image.

\textcolor{black}{
As explained in more detail in Section~\ref{sec:phisical}, the raw signal of pCLE images is acquired from tens of thousands of fibres irregularly placed in a bundle. Moreover, the LR images are reconstructed using a Delaunay-based linear interpolation that interpolates pixels from the centres of the fibres to a regular grid. Starting from the super-resolved pCLE images created by our framework, the proposed physical constraints impose that the values obtained by inverting the aforementioned interpolation are similar to the raw signal acquired from the fibres for the corresponding LR image.}

\par
To the best of our knowledge, this paper is the first to propose an adversarial network that takes advantage of the knowledge of the physical acquisition process by imposing a cycle consistency to perform unsupervised SR of medical images.
In our experiment, we show that the proposed framework does not require paired aligned patches for the training. This is an important property in all the domains where HR images are not available.
The rest of the paper is organized as follows: Section~\ref{Related} presents the state-of-the-art for unsupervised SR methods. Section~\ref{Method} presents the proposed training methodology based on an adversarial training with cycle consistency. Section~\ref{Results} presents the results obtained using a quantitative image quality assessment and a Mean Opinion Score (MOS) study and Section~\ref{Conclusion} summarizes the contribution of this research.

\section{Related work}
\label{Related}
With the recent outbreak of deep learning, example-based super-resolution (EBSR) has led to a dramatic leap in SR performance. These approaches are mainly based on a supervised training procedure where a database of aligned pairs of LR and HR images is required to create the model.
Being supervised, these SR methods are restricted to
specific training data, where the LR images are usually predetermined from their HR counterparts. However, in many contexts, such as in endomicroscopy, HR images are not available due to physical constraints and therefore these paired aligned images cannot be generated. A first attempt to train an EBSR network for endomicroscopy was proposed by~\citep{ravi} where a video-registration technique is used to estimate the HR images from a sequence of LR images. A pipeline for generation of synthetic data is finally presented to produce the desired aligned pairs. Although models trained with generated synthetic data can obtain convincing SR images, the domain gap between synthetic LR images and original pCLE images raises questions about their reliability for clinical use. For this reason, we believe that unsupervised super-resolution techniques would be more suitable in these cases. In~\citep{ayasso2012variational} is presented an unsupervised method for image SR based on a Variational Bayesian (VB) algorithm that combines a Bayesian technique with a Markovian model. The main issue with this approach is the difficulty to hand-craft a good perceptual loss function and the final images tend to be blurred. Rather than designing a suitable similarity loss function,~\citep{goodfellow2014generative} proposed a general framework called Generative Adversarial Network (GAN) where the perceptual loss function is trained directly using a discriminative network. This allows the method to automatically verify if a generated sample is similar to a real one from the target domain. In particular, the adversarial process uses two models: i) a generative model $G$, and ii) a discriminative model $D$ that are trained to play a zero-sum game. Following this general framework,~\citep{ledig2016photo} proposed a single image super-resolution architecture called SRGAN. Although this approach is unsupervised, part of its loss is still supervised. In fact, a content loss term based on a per-pixel loss between the output and ground-truth images is used there. This term requires again alignment between LR and HR thereby limiting its applicability in our context. Another drawback of SRGAN is its difficulty to train, often generating SR images that are too sharp or have artefacts. To reduce these drawbacks,~\citep{bao2017cvae} proposed to combine a VB approach with GAN. They show that an asymmetric loss function obtained using a cross-entropy loss for the discriminative network and a mean discrepancy objective for the generative network, make the GAN training more stable. Similarly to this idea, an Adversarial Variational Bayes was proposed by~\citep{mescheder2017adversarial} where a Variational Autoencoder (VAE) is trained using an auxiliary discriminative network. Contrary to the previous case, this approach provides a more clear theoretical justification. However, the problem of using paired LR/HR has not been resolved by any of the approaches described so far. One of the first approaches that formalize the possibility to translate images from a source domain $X$ to a target domain $Y$ in the absence of paired examples was proposed by~\citep{zhu2017unpaired} and is called CycleGAN. Using an adversarial training the goal of this method is to learn a mapping $G: X\rightarrow Y$ such that the distribution of images from $G(X)$ is indistinguishable from the distribution $Y$. Since this mapping is highly under-constrained, the authors also introduced an inverse mapping $F: Y\rightarrow X$ and a cycle consistency loss to ensure $F(G(X)) \approx X$. Thanks to this two-step consistency, the need for paired images is eliminated. Varying the input-output domain, this framework can be used to perform artistic style transformation~\citep{johnson2016perceptual} (where, for example, horses can be converted into zebras) or, as in our case, transfer the resolution from one domain to another. 

Some other interesting approaches were proposed by~\citep{shocher2017zero} and~\citep{bulat2018learn}. Here the authors question that the predetermined LR images obtained from standard bi-cubic down-sampling rarely look like the real LR images. In~\citep{shocher2017zero} the authors introduce a method called Zero-Shot SR, that does not rely on prior training. To do so they exploit the internal recurrence of information inside a single image and train a small image-specific CNN at test time. This facilitates self-training SR for biological data, old photos, noisy images, and other images where the acquisition process is unknown.

\textcolor{black}{Following the CycleGAN concept of~\citep{zhu2017unpaired}, we propose an unsupervised framework that uses unpaired images and is designed to overcome the limitations that standard SR approaches have when aligned pairs of LR/HR images are required.}

\section{Materials and methods}
\label{Method}
 
\begin{figure*}
  \begin{center}
     \includegraphics[width=0.80\textwidth,clip]{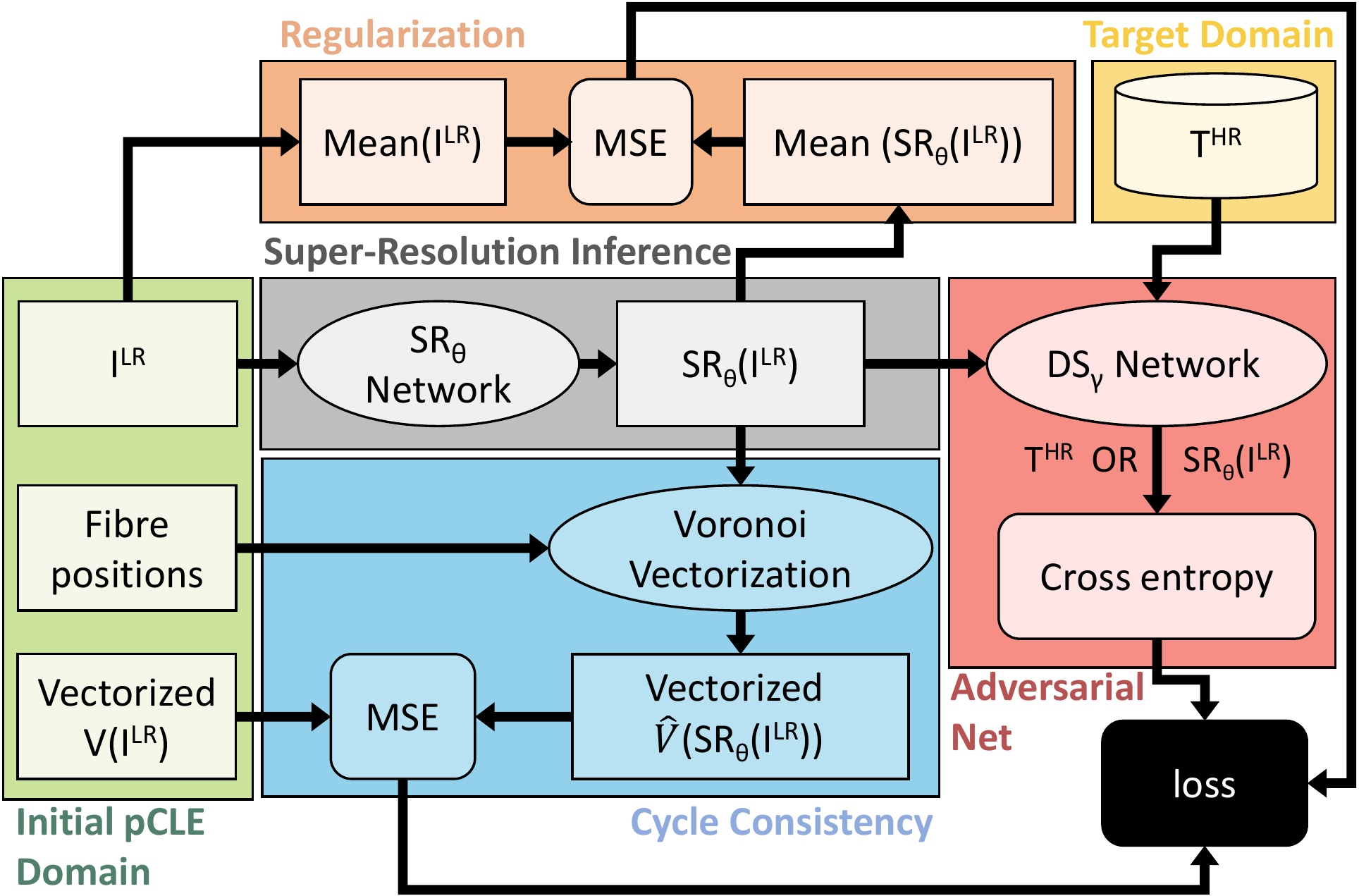}
    \caption{Pipeline used for training the proposed adversarial network with cycle consistency. Each component of the pipeline is identified by a different color.}
    \label{fig:SuperResolutionTrainingPipeline}      
\end{center}
\end{figure*}

\subsection{Database}
To validate our solution, we used the database proposed by~\citep{andre2011smart} containing 238 anonymized probe-based Confocal Laser Endomicroscopy (pCLE) video sequences from 143 patients captured on the colon and oesophagus regions. This database does not provide the real ground truth of the HR images and only estimated $\widehat{HR}$, computed using a time-consuming video-registration technique on the LR images are available. Video-registration may generate $\widehat{HR}$ that are not perfectly aligned with the LR and might display further reconstruction artefacts. \textcolor{black}{We define this set of data as $DB_{orig}$. A second version of this database called $DB_{syn}$ and based on the simulated framework proposed in~\citep{ravi} is also used in our experiments. In this case, the LR images are synthetically generated from the $\widehat{HR}$ and this results in paired images perfectly aligned. More specifically, the simulated framework extract fibre signals $fs$ directly from the $\widehat{HR}$ image, by averaging the $\widehat{HR}$ pixel values within the region defined by the Voronoi cell computed from the centre of the fibre's position. Moreover, to replicate realistic noise patterns on the simulated LR images, additive and multiplicative Gaussian noise is added to each fibre signal $fs$ obtaining a noisy fibre signal $nfs$. Finally, Delaunay-based linear interpolation is performed thereby leading to the final simulated LR images.}

\begin{figure*}[t!]
 \begin{center}

    \includegraphics[width=0.85\textwidth,clip]{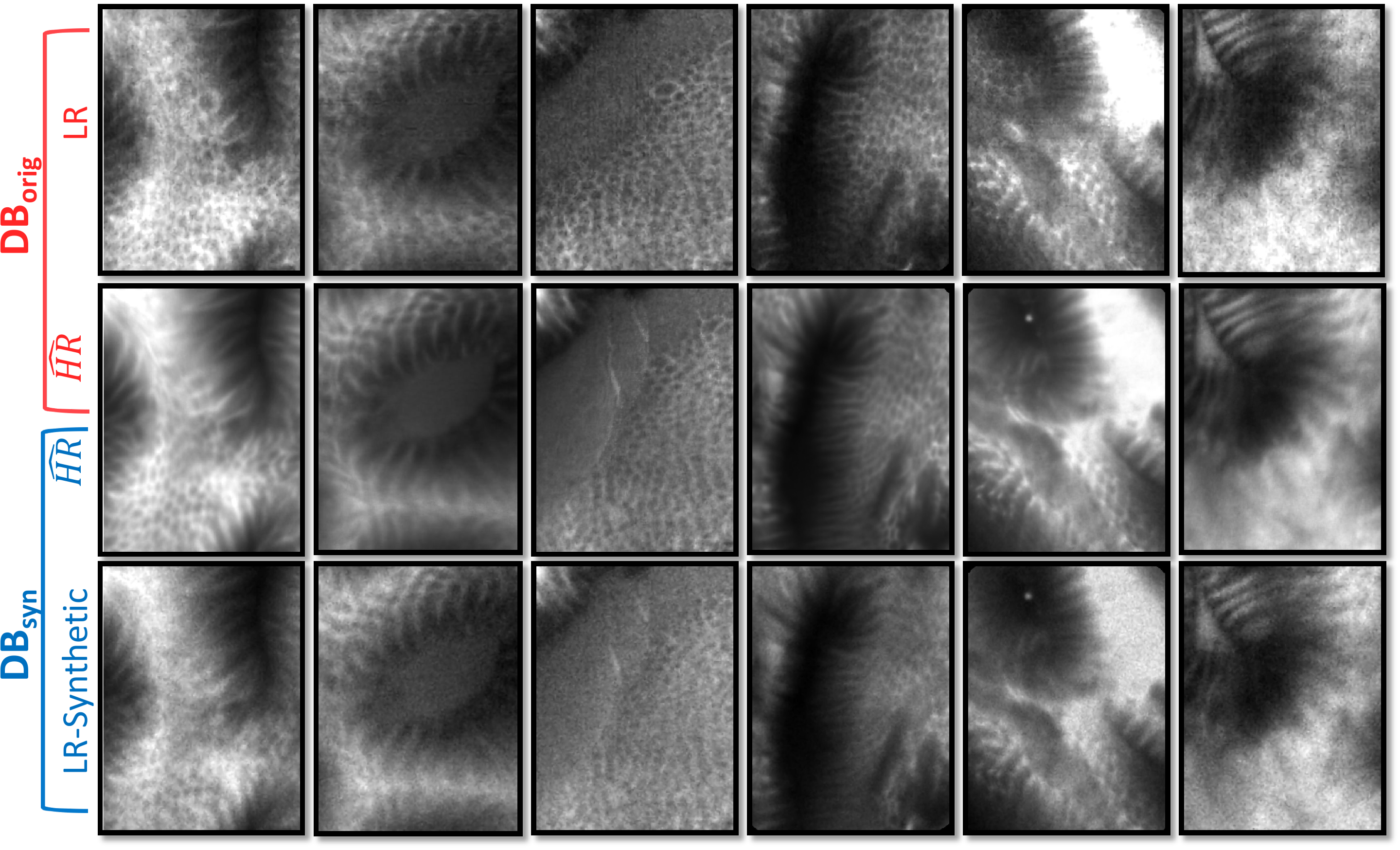}

    \caption{Example of images contained in the two proposed versions of the database. Both the two versions share the same $\widehat{HR}$ images. The LR in $DB_{orig}$ are images acquired on human tissues and they are not always aligned with the related $\widehat{HR}$. The LR in $DB_{syn}$ are instead generated synthetically and are always aligned with the corresponding $\widehat{HR}$.}
    \label{fig:Database}

\end{center}
\end{figure*}

\textcolor{black}{Fig.~\ref{fig:Database} shows some example of images extracted from both the two versions of the database. In both these scenarios, the database was divided randomly into three subsets: a train set (70\%), a validation set (15\%), and a test set (15\%). The number of images that belong to each clinical setting is maintained equal in each of these set.}
\par
\textcolor{black}{We provide the results from two different case studies: $CS_1$ where the images that belong to the same video are only distributed within one of the subsets (train, test or validation) and $CS_2$ where, additionally, the images from the same patient are only distributed within one of these subsets. $CS_1$ allows us to understand if the system is capable of super-resolve new visual patterns that have never been seen before. Given the size of our dataset, $CS_2$ allows for coarser but less prone-to-bias evaluation that mimics a more realistic scenario where the effectiveness of the system to transfer the inter-patient super-resolution capability to new patients is analysed. In our experimental section, we show that these two scenarios demonstrate similar trends.}

\textcolor{black}{Following the same pre-processed steps proposed in~\citep{ravi}, the intensity values are first normalized to have mean zero and variance one, then the pixels values were scaled of every frame individually in the range [0–1] and finally, non-overlapping patches of 64$\times$64 pixels were extracted only from the pCLE field of view of the train and validation set. The patches in the validation set were used to monitor the loss and avoid overfitting. Test images with size 512$\times$512 were processed at full-size to compute the final results. We highlight that the proposed SR framework does not have the aim to increase the number of pixels, but rather to improve the quality of the LR images that are initially oversampled with an average of 7 interpolated pixels for each informative fibre-pixel. The choice to initially oversample is made by the manufacturer to ensure that the image space is discretised in a sufficiently fine manner to map the fibre graph onto a square pixel grid without too much distortion. In conclusion, in our system, the output images have the same size as the input images but display refined content. The full-size processing of the test images is possible since the inference network is fully convolutional and no specific image size is required as input.}

\subsection{Adversarial training}
The pipeline used for training our framework is presented in Fig.~\ref{fig:SuperResolutionTrainingPipeline} and is divided into different sub-sections, each coded by a specific colour. 

We formalize our training as an adversarial min-max problem where two networks, a discriminative network defined as $DS_{\gamma}$ (red sub-sequence in Fig.~\ref{fig:SuperResolutionTrainingPipeline}), and a super-resolution network defined as $SR_{\theta}$ (grey sub-sequence in Fig.~\ref{fig:SuperResolutionTrainingPipeline}) are trained concurrently. More specifically, the first network $DS_{\gamma}$ is trained solving:

\begin{equation}\label{minmax}
\max\limits_{\gamma}\mathbb{E}_{x\sim p_{I^{LR}}}\big[ log\big(1-DS_{\gamma}\big(SR_{\theta}(x)\big)\big) \big]+\mathbb{E}_{y\sim p_{T^{HR}}}\big[ log DS_{\gamma}(y) \big],
\end{equation} where $p_{I^{LR}}$ and $p_{T^{HR}}$ are respectively the patch distributions on the input and target domain, $DS_{\gamma}(*)$ estimates the probability that a patch comes from the target domain, whereas $SR_{\theta}(x)$ is the predicted super-resolved patch obtained from $x$. The meaning of Eq.~\ref{minmax} is that the discriminator has to maximize how to discriminate predicted super-resolved images from real $T^{HR}$ patches.

The second network $SR_{\theta}$, is trained instead through the minimization of a composite loss function $loss_t$ obtained solving:
\begin{equation}
\min\limits_{\theta}\mathbb{E}_{x\sim p_{I^{LR}}}\big[loss_t \big(x,SR_{\theta}(x)\big)\big]
\end{equation}

The proposed $loss_t$, defined in Eq.~\ref{loss_t}, is a combination of three terms: $l_{Vec}$ that models the physical acquisition characteristics of the predicted super-resolved patch, $l_{Adv}$ that models the adversarial loss function and $l_{Reg}$ used to regularize the network training. The details of each term are provided later in this section.

\begin{equation}\label{loss_t}
loss_t=l_{Vec}+l_{Adv}+l_{Reg}
\end{equation}

Both $SR_{\theta}$ and $DS_{\gamma}$ are concurrently trained using the back-propagation algorithm that gradually adjusts the parameters $\theta$ and $\gamma$ through a stochastic gradient descent for the former and a stochastic gradient ascent for the latter.

\subsection{Input domain and cycle consistency}\label{sec:phisical}
\subsubsection{Input domain}
The green blocks in Fig.~\ref{fig:SuperResolutionTrainingPipeline} represent the data structures required as input for the proposed pipeline.
The most obvious input is the reconstructed $I^{LR}$ that is used by the $SR_{\theta}$ network to infer the super-resolved patch $SR_{\theta}(I^{LR})$.

In the pCLE imaging, image acquisition is achieved by illuminating one fibre at a time. Each fibre acts as an individual pinhole and a scan point for fibre confocality. The information from all the fibres is then collected in a vector that we refer to as a vectorized image $V(I^{LR})$, and represents the main input block in our pipeline. $I^{LR}$ images are reconstructed interpolating the values in $V(I^{LR})$ from the centres of the fibre positions to the points of a regular grid. Therefore the fibre positions are the other key input block required by our pipeline. 
\subsubsection{Cycle consistency}
\label{Cycleconsistency}

\begin{figure*}[t!]
  \begin{center}
     \includegraphics[width=0.95\textwidth,clip]{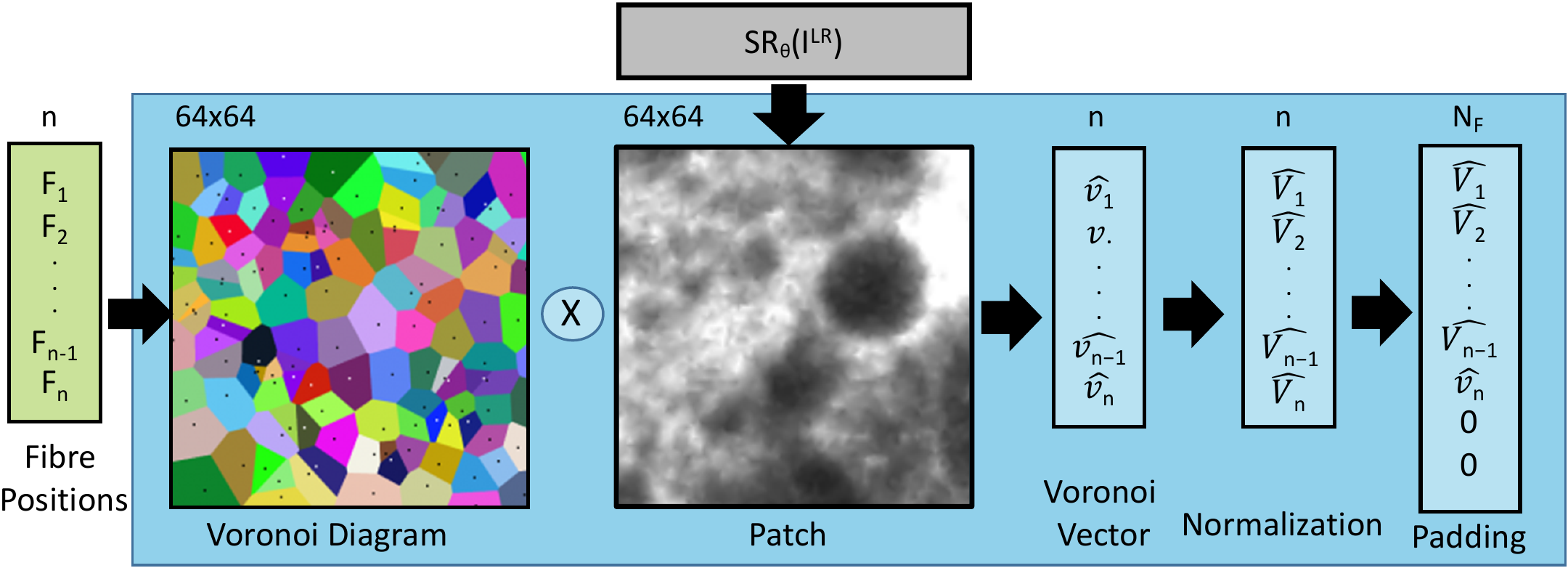}
    \caption{Voronoi vectorization used in our pipeline to constrain the predicted super-resolved patches.}
    \label{fig:VoronoiVectorizat}      
\end{center}
\end{figure*}

Starting from a generated high-resolution pCLE image $SR_{\theta}(I^{LR})$, we can obtain a low-resolution representation of it, by a process referred to as Voronoi vectorization $\widehat{V}(SR_{\theta}(I^{LR}))$ which is equivalent to the down-sampling for standard images.
The details of the Voronoi vectorization used in our framework are described in Fig.~\ref{fig:VoronoiVectorizat}. Here, the first step is to compute the Voronoi diagram from the fibre positions. The result is a partition of the plane where for each fibre there is a corresponding region, called Voronoi cell, consisting of all points closer to this fibre than to any other fibre. The next step is to average the pixels in the $SR_{\theta}(I^{LR})$ patch that belongs to the same Voronoi cell, imitating the point spread function of the fibre acquisition process. All the elements in the vector are then normalized in the range [0, 1]. This normalization makes the training faster and reduces the chances of getting stuck in local optima. Since each patch may have a different number of fibres, the vectorization can produce vectors of different sizes. \textcolor{black}{Therefore as a final step, a 0-padding is introduced so that each vector always has a fixed number of elements. We define this fixed number as $N_F$ that is equal to the maximum number of fibres in a single patch. In our database $N_F$ is 682 which is commensurate with the ratio between the patch size (64$\times$64) and the average factor (7) used to oversample each informative fibre-pixel.}

The vectorized $V(I^{LR})$ and the Voronoi vectorization $\widehat{V}(SR_{\theta}(I^{LR}))$ are used in our pipeline to create the cycle consistency (blocks coloured in cyan in Fig.~\ref{fig:SuperResolutionTrainingPipeline}). These blocks are used to impose the requirements for the predicted super-resolved images $SR_{\theta}(I^{LR})$ to have the same physical acquisition properties as the initial $I^{LR}$ images. Without this cycle consistency, the network could simply produce arbitrary images in the target domain with no relationship to the structures contained in the input image, because our framework relies on unpaired patches. To avoid this, we force the $V(I^{LR})$ and $\widehat{V}(SR_{\theta}(I^{LR}))$ to be similar using the $l_{Vec}$ term in the proposed loss function.

\begin{equation}
l_{Vec}=\frac{1}{N_F}\sum_{i=1}^{N_F}\bigg[V(I^{LR})_i-\widehat{V}(SR_{\theta}(I^{LR}))_i\bigg]^2
\end{equation}

In contrast to CycleGan~\citep{zhu2017unpaired}, our cycle consistency block is not a trainable network, but rather is used to constrain the $SR_{\theta}$ network to generate images with the same physical acquisition properties as the initial $I^{LR}$ images.

\subsection{Super-resolution network}
We decided to use the layout for the SR network proposed in~\citep{ledig2016photo}. $SR_{\theta}$ is aimed at producing images that are similar to the one in the target domain by trying to fool the discriminator network. This is achieved through the term $l_{Adv}$ in the proposed loss function defined as follows:

\begin{equation}
l_{Adv}=-log DS_{\gamma}(SR_{\theta}(I^{LR})),
\end{equation} where $DS_{\gamma}(SR_{\theta}(I^{LR}))$ is the probability that the predicted image $SR_{\theta}(I^{LR})$ is classified as a real $T^{HR}$. As proposed by~\citep{goodfellow2014generative} we minimize $-log DS_{\gamma}(SR_{\theta}(I^{LR}))$ instead
of $log[1- DS_{\gamma}(SR_{\theta}(I^{LR}))]$ for better
gradient behavior.

In the inference phase, only $SR_{\theta}$ is used for processing the $I^{LR}$ images.

\subsection{Regularization}
The blocks displayed in orange in Fig.~\ref{fig:SuperResolutionTrainingPipeline} are used to regularize the network training. This regularization is required since the Voronoi vectorization of each patch is normalized to the range [0-1] and this may result in an expansion of its histogram range. To restore the correct histogram distribution, we impose that the mean values in each row and each column of the patch are identical between the initial $I^{LR}$ and the obtained $T^{HR}$. This is achieved in our framework through the $l_{Reg}$ term of $loss_t$: 

\footnotesize
\begin{equation}
l_{Reg}= \frac{1}{H}\sum_{y=1}^{H}\Bigg[\frac{1}{W}\sum_{x=1}^W SR_{\theta}(I^{LR}_{xy})-\frac{1}{W}\sum_{x=1}^W I^{LR}_{xy}\Bigg]^2+
\frac{1}{W}\sum_{x=1}^{W}\Bigg[\frac{1}{H}\sum_{y=1}^H SR_{\theta}(I^{LR}_{xy})-\frac{1}{H}\sum_{y=1}^H I^{LR}_{xy}\Bigg]^2
\end{equation}

\normalsize

\subsection{Training domain}
\textcolor{black}{In our pipeline we considered four different target domains to transfer the super-resolution to the initial LR images: i) $T^{HR}_{nat}$ where the HR patches are extracted from natural images (grey-scaled images from the Sun2012 database~\citep{xiao2010sun}), ii) $T^{HR}_{orig}$ containing the HR patches obtained by the video-registration technique on the LR images, iii) $T^{HR}_{syn}$ containing paired HR patches obtained using the video-registration technique while the LR are synthetically aligned, 
and iv) $T^{HR}_{res}$ where the HR patches are obtained by down-sampling large portions of the LR images by a factor of four. Inspired by the work proposed in~\citep{shocher2017zero}, the idea behind this last target domain is based on the fact that patches in the images have recurrences at a different scale and down-sampling large LR images may increase the high-frequency responses in the generated down-sampled HR patches.}

\subsection{Training details and parameters}
In our implementation, Eq.~\ref{minmax} is solved by minimizing the cross-entropy of the number of samples correctly discriminated by $DS$. As proposed by~\citep{arjovsky2017towards} we add white noise to the inputs of the $DS_{\gamma}$ network to stabilize the adversarial training. We trained our networks on an NVIDIA GTX TITAN-X GPU card with 12 GB of memory. The training procedure converges after 50-80 thousand iterations of random mini-batch with 54 patches. For the optimization of the stochastic gradient descent, we use Adam with $\beta 1 = 0.9$, $\beta 2 = 0.999$ and $\epsilon =$ 10e-8. The networks were trained with a learning rate of 10e-4.
\section{Experiments}

\label{Results}
\begin{figure*}[t!]
  \begin{center}

    \includegraphics[width=0.95\textwidth,clip]{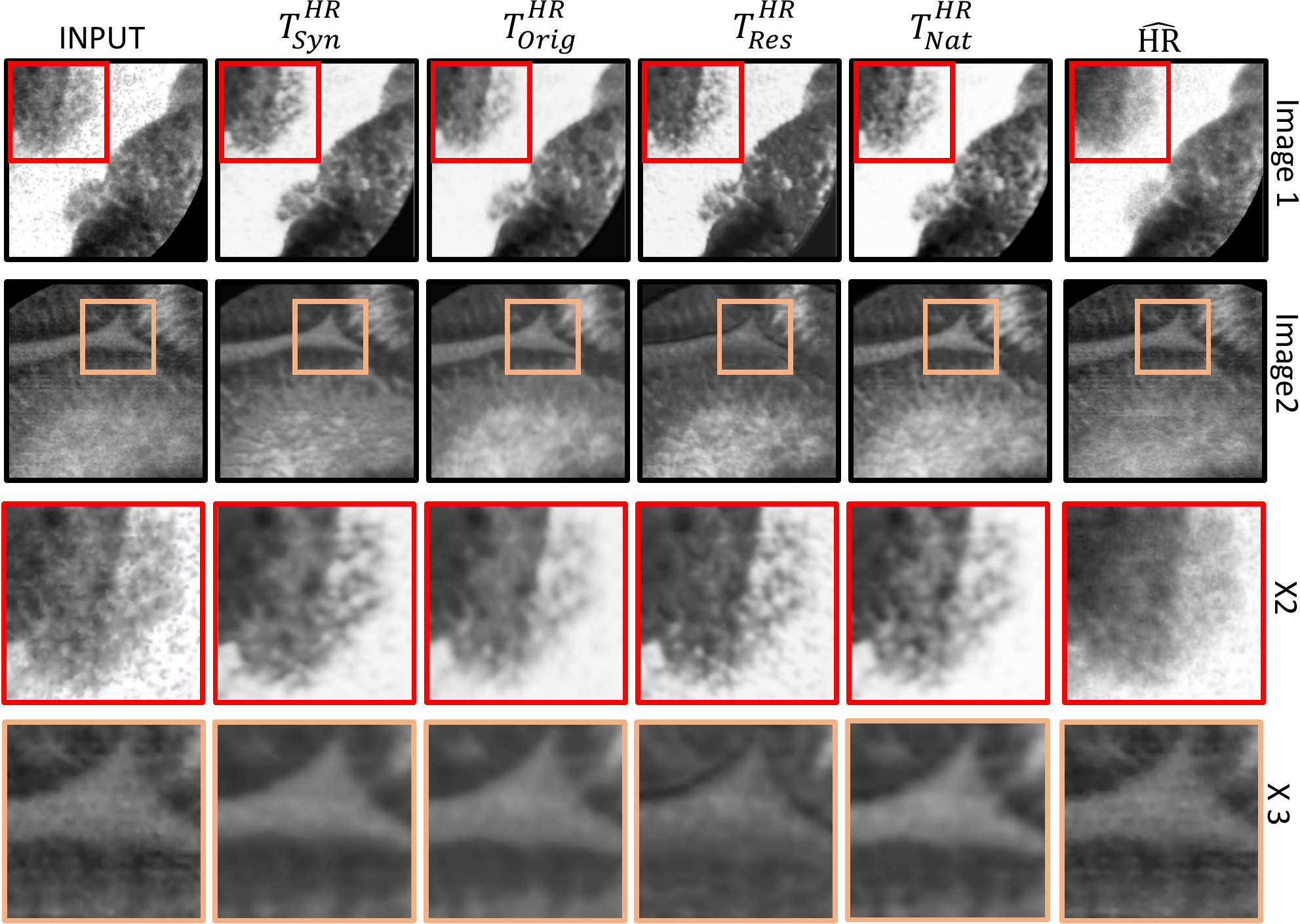}

    \caption{Example of visual results obtained by the proposed approaches when trained with different target domains. From left to right we have: Input, training with $T^{HR}_{syn}$, training with $T^{HR}_{orig}$, training with $T^{HR}_{res}$, training with $T^{HR}_{nat}$ and $\widehat{HR}$.}
    \label{fig:SuperResolutionDiffentConfigurations}    

\end{center}
\end{figure*}

Due to the lack of real ground truth in our database, the validation of our experiments is based on complementary quantitative and qualitative analysis. The quantitative analysis, presented in Section~\ref{QuantitativeAnalisi}, uses four different metrics to evaluate the obtained images. The qualitative analysis is instead based on a MOS study carried out by clinicians and medical imaging experts that gave numerical indications of the perceived quality of the super-resolved images.

\subsection{Quantitative analysis}
\label{QuantitativeAnalisi}
The four metrics used in our quantitative analysis are: i) a Structural Similarity matrix (SSIM) proposed by~\citep{wang2004image} that evaluates the similarity between $SR_{\theta}(I^{LR})$ and $\widehat{HR}$, ii) $\Delta GCF_{ \widehat{HR}}$ that quantifies the improvement on the global contrast factor (a reference-free metric for measuring image contrast~\citep{matkovic2005global}) that the super-resolved image yields with respect to $\widehat{HR}$, iii) $\Delta GCF_{I^{LR}}$ that is the improvement of the global contrast factor that the super-resolved image yields with respect to the initial $I^{LR}$, and iv) a composite score $Tot_{cs}$ obtained by normalizing the value of SSIM and $\Delta GCF_{ \widehat{HR}}$ in the range [0,1] and averaging the obtained results. \textcolor{black}{The formula used to compute $Tot_{cs}$ is described by Eq.~\ref{TotCs}}.

\begin{equation}\label{TotCs}
Tot_{cs}=\frac{\frac{SSIM_{\widehat{HR}}-0.6}{0.4}+\frac{\Delta GCF_{\widehat{HR}}+0.5}{1.8}}{2}
\end{equation}

This composite score leads to a more robust evaluation of the results since, SSIM alone is not reliable when the ground truth is only estimated, while the GCF can be improved by merely adding random high frequency to the images.

\begin{table*}[t!]
\caption{Quantitative analysis results obtained by our approach when trained with different target domains on case study $CS_1$ and $CS_2$.}
\label{table:Configuration}
\begin{center}
 
\resizebox{0.80\textwidth}{!}{%
\begin{tabular}{c|c|>{\centering\arraybackslash}p{2.5cm}>{\centering\arraybackslash}p{2.5cm}>{\centering\arraybackslash}p{2.5cm}>{\centering\arraybackslash}p{2.5cm}}
    \multirow{7}{*}{\rotatebox{90}{$CS_1$}}&&$T^{HR}_{syn}$&    $T^{HR}_{orig}$&    $T^{HR}_{res}$&   $T^{HR}_{nat}$\\
    \hline
&$SSIM_{\widehat{HR}}$& 0.90 $\pm$ 0.03 & \textbf{0.91 $\pm$ 0.03} & 0.87 $\pm$ 0.04 & 0.86 $\pm$ 0.03 \\
&$\Delta GCF_{ \widehat{HR}}$& 0.01 $\pm$ 0.29 & 0.38 $\pm$ 0.27 & -0.13 $\pm$ 0.40 & \textbf{0.66 $\pm$ 0.31} \\
&$\Delta GCF_{I^{LR}}$& -0.28 $\pm$ 0.19 & 0.09 $\pm$ 0.19 & -0.42 $\pm$ 0.31  & \textbf{0.38 $\pm$ 0.26} \\
\cline{2-6} 
&$Tot_{cs}$& 0.52 & 0.63 & 0.44 & \textbf{0.64} \\
\specialrule{2.5pt}{1pt}{1pt}

\multirow{ 4}{*}{\rotatebox{90}{$CS_2$}}
&$SSIM_{\widehat{HR}}$& \textbf{0.91 $\pm$ 0.03} & \textbf{0.91 $\pm$ 0.03} & 0.87 $\pm$ 0.03 & 0.86 $\pm$ 0.04 \\
&$\Delta GCF_{ \widehat{HR}}$& -0.10 $\pm$ 0.36 & 0.24 $\pm$ 0.35 & -0.26 $\pm$ 0.40 & \textbf{0.51 $\pm$ 0.40} \\
&$\Delta GCF_{I^{LR}}$& -0.33 $\pm$ 0.21 & 0.01 $\pm$ 0.18 & -0.49 $\pm$ 0.28  & \textbf{0.28 $\pm$ 0.25} \\
\cline{2-6} 
&$Tot_{cs}$& 0.49 & 0.59 & 0.41 & \textbf{0.61} \\

\end{tabular}
}
\end{center}
\end{table*}

\textcolor{black}{Our first experiment is aimed at finding the best target domain for improving the pCLE images. The results computed on $DB_{orig}$ for both the case studies are reported in Table~\ref{table:Configuration}. As we can see, the network trained with natural images ($T^{HR}_{nat}$) obtains the best $Tot_{cs}$ score. 
From these results, we can also deduct that using synthetic images for the training is worse than using images from the original domain. This is probably due to the fact that synthetic images may have a non negligible domain gap with the real images. With this result, we can state that paired patches are not anymore a requirement for our framework. Finally, downsampling LR images to create patches with higher frequency content does not seem to provide good results and the high-frequency signals are not recovered. These qualitative indications can be seen on reconstructed images reported in Fig.~\ref{fig:SuperResolutionDiffentConfigurations}.
}
\par
\textcolor{black}{Looking at the different case studies, the aforementioned considerations are consistent along both the cases, although $CS_2$ shows slightly lower performances with respect to $CS_1$ probably due to the fact that it has a coarser split of its dataset.}

\begin{table*}[t!]
\caption{Quantitative analysis results of the proposed approach against state-of-the-art methods on the database $DB_{orig}$ for case study $CS_1$ and $CS_2$.}
\label{table:StateOfTheArt1}
\begin{center}
\resizebox{1\textwidth}{!}{%

\begin{tabular}{c|c|>{\centering\arraybackslash}p{3cm}>{\centering\arraybackslash}p{2.5cm}>{\centering\arraybackslash}p{2.5cm}>{\centering\arraybackslash}p{2.5cm}>{\centering\arraybackslash}p{2.5cm}}
    \multirow{ 8}{*}{\rotatebox{90}{$CS_1$}}&&Proposed& Ravi-Szcz. [2018]& [Villena 2013] & Wiener&    Contrast-enhancement    \\
\hline
&$SSIM_{\widehat{HR}}$& 0.86 $\pm$ 0.03 & \textbf{0.88 $\pm$ 0.05} & 0.86 $\pm$ 0.05 & 0.83 $\pm$ 0.08 & 0.62 $\pm$ 0.08 \\
&$\Delta GCF_{ \widehat{HR}}$& 0.66 $\pm$ 0.31 & 0.42 $\pm$ 0.24 & 0.27 $\pm$ 0.23 & -0.00 $\pm$ 0.37 & \textbf{1.34 $\pm$ 0.36} \\
&$\Delta GCF_{I^{LR}}$& 0.38 $\pm$ 0.26 & 0.13 $\pm$ 0.13 & -0.02 $\pm$ 0.07 & -0.29 $\pm$ 0.27 &\textbf{ 1.06 $\pm$ 0.25} \\
\cline{2-7} 
&$Tot_{cs}$& \textbf{0.64} & 0.61 & 0.54 & 0.42 & 0.53 \\

\specialrule{2.5pt}{1pt}{1pt}

\multirow{ 4}{*}{\rotatebox{90}{$CS_2$}}
&$SSIM_{\widehat{HR}}$& 0.86 $\pm$ 0.04 & \textbf{0.89 $\pm$ 0.04} & 0.88 $\pm$ 0.04 & 0.85 $\pm$ 0.06 & 0.63 $\pm$ 0.06 \\
&$\Delta GCF_{ \widehat{HR}}$& 0.51 $\pm$ 0.40 & 0.38 $\pm$ 0.29 & 0.21 $\pm$ 0.29 & -0.15 $\pm$ 0.44 & \textbf{1.32 $\pm$ 0.34} \\
&$\Delta GCF_{I^{LR}}$& 0.28 $\pm$ 0.25 & 0.15 $\pm$ 0.09 & -0.02 $\pm$ 0.09 & -0.38 $\pm$ 0.31 &\textbf{ 1.09 $\pm$ 0.18} \\
\cline{2-7} 
&$Tot_{cs}$& \textbf{0.61} & 0.60 & 0.54 & 0.41 & 0.54 \\

\end{tabular}
}
\end{center}
\end{table*}

\begin{table*}[t!]
\caption{Quantitative analysis results of the proposed approach against state-of-the-art methods on the database $DB_{syn}$ for case study $CS_1$ and $CS_2$.}
\label{table:StateOfTheArt2}
\begin{center}
\resizebox{1\textwidth}{!}{%

\begin{tabular}{c|c|>{\centering\arraybackslash}p{3cm}>{\centering\arraybackslash}p{2.5cm}>{\centering\arraybackslash}p{2.5cm}>{\centering\arraybackslash}p{2.5cm}>{\centering\arraybackslash}p{2.5cm}}
     \multirow{ 8}{*}{\rotatebox{90}{$CS_1$}}&&Proposed& Ravi-Szcz. [2018]& [Villena 2013] & Wiener&    Contrast-enhancement    \\
    \hline
&$SSIM_{\widehat{HR}}$& 0.90 $\pm$ 0.03 & \textbf{0.93 $\pm$ 0.03} & 0.89 $\pm$ 0.05 & 0.88 $\pm$ 0.06 & 0.66 $\pm$ 0.08 \\
&$\Delta GCF_{ \widehat{HR}}$& 0.60 $\pm$ 0.22 & 0.45 $\pm$ 0.25 & 0.13 $\pm$ 0.11 & -0.21 $\pm$ 0.21 &\textbf{ 1.03 $\pm$ 0.32} \\
&$\Delta GCF_{I^{LR}}$& 0.47 $\pm$ 0.26 & 0.32 $\pm$ 0.32 & 0.00 $\pm$ 0.00 & -0.34 $\pm$ 0.19 &\textbf{ 0.90 $\pm$ 0.26 }\\
\cline{2-7} 
&$Tot_{cs}$& \textbf{0.68} & \textbf{0.68} & 0.54 & 0.43 & 0.50\\

\specialrule{2.5pt}{1pt}{1pt}

\multirow{ 4}{*}{\rotatebox{90}{$CS_2$}}
&$SSIM_{\widehat{HR}}$& 0.91 $\pm$ 0.03 &\textbf{ 0.92 $\pm$ 0.03} & 0.90 $\pm$ 0.04 & 0.89 $\pm$ 0.06 & 0.65 $\pm$ 0.07 \\
&$\Delta GCF_{ \widehat{HR}}$& 0.50 $\pm$ 0.25 & 0.52 $\pm$ 0.20 & 0.11 $\pm$ 0.05 & -0.28 $\pm$ 0.22 & \textbf{1.08 $\pm$ 0.20} \\
&$\Delta GCF_{I^{LR}}$& 0.39 $\pm$ 0.23 & 0.41 $\pm$ 0.23 & -0.00 $\pm$ 0.00 & -0.38 $\pm$ 0.21 & \textbf{0.97 $\pm$ 0.18 }\\
\cline{2-7} 
&$Tot_{cs}$& 0.66 & \textbf{0.68} & 0.54 & 0.42 & 0.50 \\

\end{tabular}
}
\end{center}
\end{table*}

\begin{figure*}[t!]
\begin{center}
  \includegraphics[width=0.95\textwidth,clip]{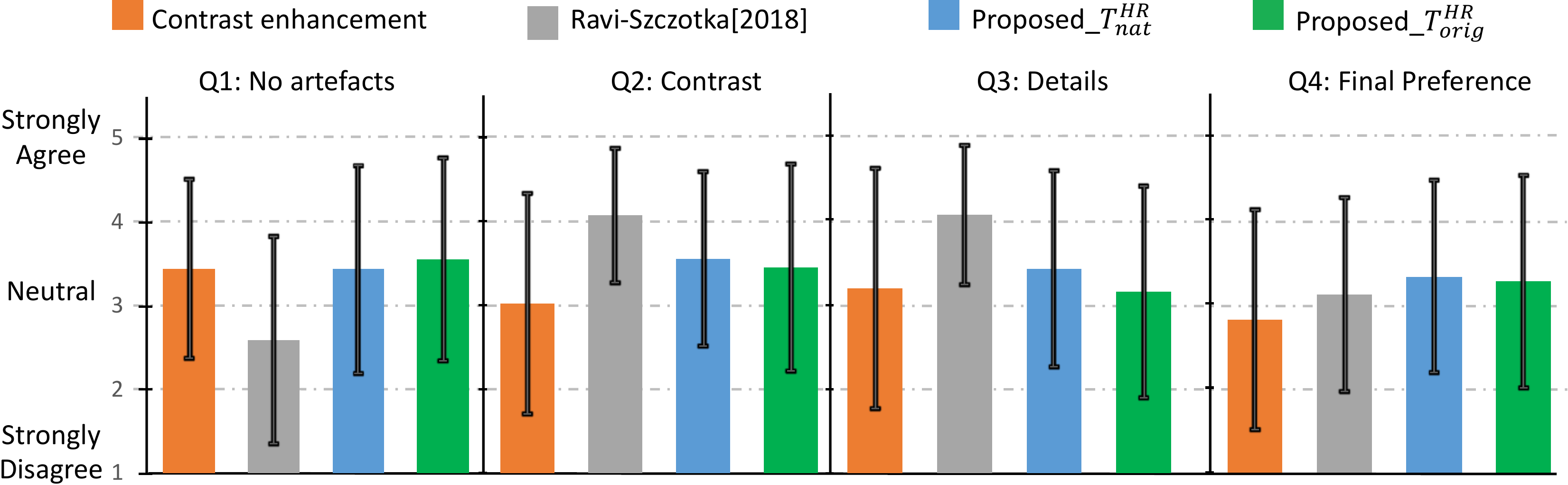}
\caption{Mean and standard deviation of the participants' replies to each of the four MOS questions for the evaluation of the results obtained by contrast-enhancement (baseline),~\citep{ravi}, and proposed approach trained using two different target domains ($T^{HR}_{nat}$ and $T^{HR}_{orig}$). These results were obtained on the test images of $DB_{orig}$ for the case study $CS_1$.}
\label{fig:MOSResults}       
\end{center}
\end{figure*}

To further validate our framework, we compare our best approach \textcolor{black}{(the network trained with the target domain $T^{HR}_{nat}$)}, against some state-of-the-art single image super-resolution methodologies. These results are presented in Table~\ref{table:StateOfTheArt1} for the database $DB_{orig}$ and in Table~\ref{table:StateOfTheArt2} for the database $DB_{syn}$. In these experiments we consider three different approaches: i) the unsupervised Wiener deconvolution tuned on the train set, ii) the unsupervised variational Bayesian inference approach with sparse and non-sparse priors~\citep{villena2013bayesian}, and the supervised EBSR proposed by~\citep{ravi}. Finally, a contrast-enhancement approach obtained by sharpening the input was also used as a baseline.
\par
\textcolor{black}{In the dataset $DB_{orig}$, although the sharpening algorithm produces the best contrast improvements, our approach obtains the highest SSIM and, according to $Tot_{cs}$, the overall performance outperforms all the other approaches.}

\textcolor{black}{Differently from the results obtained with $DB_{orig}$, with the database $DB_{syn}$, our approach is not able to overcome the results obtained by~\citep{ravi}. This is probably due to the fact that the supervised training in~\citep{ravi} exploits principles that are similar to the ones used to generate the synthetic images in $DB_{syn}$. Therefore, the results obtained by this approach in the database $DB_{syn}$ are obtained in a purely simulated scenario where the model is trained on data with no domain gap with the test set. Consequently, the supervised approach in~\citep{ravi} has an advantage with respect to our unsupervised one in this setting. What is interesting to see nonetheless, is that our unsupervised framework can achieve almost similar performance to the supervised one of~\citep{ravi} despite the evaluation being intrinsically favourable for this last solution.}
\par
\textcolor{black}{Also in these two experiments, close results are obtained between $CS_1$ and $CS_2$ confirming that the system is able to super-resolve images for both the considered cases (i.e. when the images contain new visual structures or when they are extracted from new patients).}

\textcolor{black}{The statistical significances of the improvements discussed in this section were assessed with a paired t-test and the p-values are all less than 0.0001.}

\subsection{Semi-quantitative analysis (MOS)}
\label{MOS}
\textcolor{black}{To perform the MOS, we asked 10 trained individuals to evaluate, on average, 20 images each, randomly selected from the test set of $DB_{orig}$ on the case study $CS_1$. At each step, the SR images obtained with two different configurations of the proposed approach, with~\citep{ravi}, and with a contrast-enhancement approach that sharpens the input (baseline), were shown to the user in a random order to reduce any possible bias on the evaluation of the images. The two configurations used for our approach were the one based on training our model with the two best target domains (i.e. $T^{HR}_{nat}$ and $T^{HR}_{orig}$).} The input and the $\widehat{HR}$ were also displayed on the screen as references for the participants. For each of the four images, the user assigned a score between 1 (strongly disagree) to 5 (strongly agree) on the following questions:
\small
\begin{itemize}
    \item \textit{Q1: Is the image artefact-free?}
    \item \textit{Q2: Can you see an improvement in contrast with respect to the input?}
    \item \textit{Q3: Can you see an improvement in the details with respect to the input?} 
    \item \textit{Q4: Would you prefer seeing the new image over the input?} 
\end{itemize}
\normalsize

To make sure that the questions were consistently interpreted, each participant received a short training before starting the study. The results on the MOS presented in Fig.~\ref{fig:MOSResults} show that, \textcolor{black}{between the two different configurations used on our approach, the model trained with natural images $T^{HR}_{nat}$ provides a better trade-off of artefacts, contrast and details with respect to the training using the target domain $T^{HR}_{orig}$. The results show also that the proposed approach and~\citep{ravi} provide complementary features. In fact, although the details (question Q3) and the contrast (question Q2) in both the settings of our approach seem to be worse than~\citep{ravi}, our solution provides better scores for the absence of artefacts (question Q1), which is an important characteristic in clinical applications. Regarding the final preference (question Q4), our solution trained using natural images ($T^{HR}_{nat}$) shows the best results with respect to all the other approaches, confirming the validity of our solution to perform super-resolution on pCLE images}. The approach that sharpens the images is, instead, the one that provides the lowest scores for Q2, Q3 and Q4, probably because it enhances the noise.

\begin{figure*}[t!]
  \begin{center}

    \includegraphics[width=0.95\textwidth,clip]{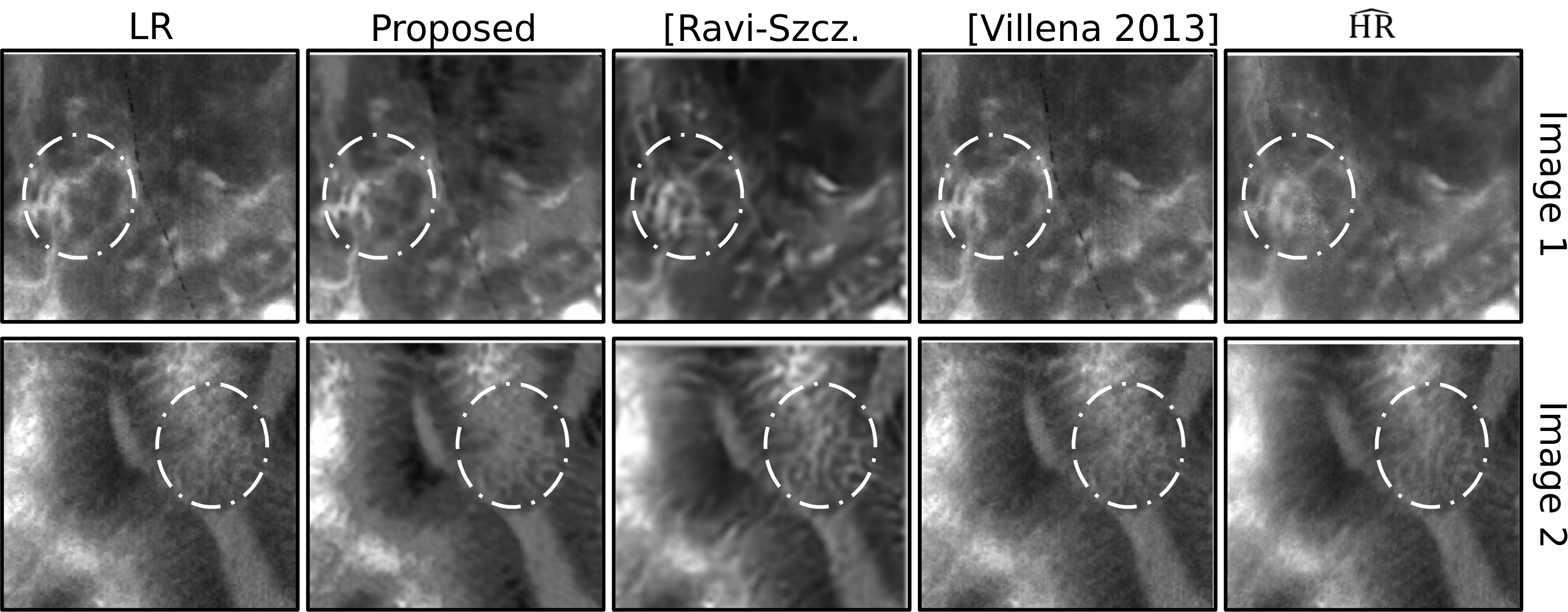}

    \caption{Example of visual results obtained by our approach in comparison with other state-of-the-art approaches. From left to right we have: Input image, proposed output, output from~\citep{ravi}, output from~\citep{villena2013bayesian} and $\widehat{HR}$.}
    \label{fig:SuperResolutionExamples}

\end{center}
\end{figure*}

Visual results for some of these images are shown in Fig.~\ref{fig:SuperResolutionExamples}, confirming these findings. More specifically, although the output proposed by~\citep{ravi} shows better contrast and higher sharpness images, the corresponding algorithm also behaves much worse in term of artefacts generation. In fact, as we can see in Fig.~\ref{fig:SuperResolutionExamples}, it often enhances noise and make up details that are not visible neither in the input nor in the estimated HR images and this can eventually lead to a wrong clinical interpretation of the images. In Fig.~\ref{fig:SuperResolutionExamples} we have marked with a white circle some of the regions where these issues are more evident. From a clinical point of view, we believe that the reliability of SR images in terms of details is a more important feature than having high contrast or high sharpness created by artefacts. According to these visual considerations and the MOS findings, we can conclude that our solution provides a more convincing representation for super-resolved pCLE images with respect to the other state-of-the-art approaches.

\section{Discussion and conclusions}
\label{Conclusion}
Obtaining medical images that accurately visualize structures of tissues is still today an open challenge. One of the main issue that researchers are trying to address here, is to improve the image resolution. In endomicroscopy, low image resolution is often dependent on the intrinsic limitations of the acquisition systems. Current solutions propose SR methods to post-process the final images as an alternative to the more difficult hardware enhancements. Clinical impacts and benefits in the use of SR methods include: i) better localization of tissue structures, ii) improving the image contrast and iii) improving the Signal to Noise Ratio (SNR)~\citep{greenspan2008super}. However, often the validation of these benefits in terms of clinical outcome is not straightforward. An attempt for this was proposed in~\citep{kennedy2006super}, where Positron Emission Tomography (PET) scans on phantom and patients were used to prove that smaller visual features were localized and better visualized using SR techniques than without. Another similar study conducted in~\citep{kennedy2007improved} shows that using SR techniques produces better contrast ratios and better target-to-background ratios than the standard reconstructions.~\citep{plenge2012super} designed, instead, an experimental framework to show that the SR reconstructions are more advantageous in terms of the SNR with respect to the direct HR acquisition.
Finally,~\citep{sano2017super} proposed a novel measurement algorithm for joint space distance on X-ray images generated by a SR method. The results exhibit higher accuracy in the measured distances when SR images were used.
\par
\textcolor{black}{The studies above show that SR methods can improve the clinical outcomes and can open the door for better diagnosis. In our case, however, the lack of a real ground truth can raise some scepticism on the validation of the results since is not simple to show that SR approaches don't emphasize or make up details that are not real. Our extensive quantitative and qualitative analysis, based also on expert’s evaluations, are developed to show the reliability of the obtained SR images and the reduced presence of artefacts even in the absence of real ground truth. More specifically, these results validated using two versions of a database containing 238 endomicroscopy video sequences captured from 143 patients demonstrate the ability of the pipeline to produce convincing super-resolved images.}

In conclusion, in our study we report a super-resolution framework for endomicroscopy images based on an unsupervised adversarial deep neural network that takes advantage of the knowledge of the physical acquisition process to impose a cycle consistency. The proposed framework results to be particularly useful in all situations where there is a lack of HR images and pairs of LR/HR images are not available for the supervised training.

To the best of our knowledge, we are the first to propose an unsupervised super-resolution approach for medical images. Further clinical studies could validate the relevance of the proposed framework to specific clinical applications for super-resolution.

\section*{Acknowledgement}

\textbf{Funding:} This work was supported by Wellcome/EPSRC [203145Z/16/Z; NS/A000050/1; WT101957; NS/A000027/1; EP/N027078/1]. This work was undertaken at UCL and UCLH, which receive a proportion of funding from the DoH NIHR UCLH BRC funding scheme. The PhD studentship of Agnieszka Barbara Szczotka is funded by Mauna Kea Technologies, Paris, France.


\begin{thebibliography}{25}
\providecommand{\natexlab}[1]{#1}
\providecommand{\url}[1]{\texttt{#1}}
\expandafter\ifx\csname urlstyle\endcsname\relax
  \providecommand{\doi}[1]{doi: #1}\else
  \providecommand{\doi}{doi: \begingroup \urlstyle{rm}\Url}\fi

\bibitem[Andr{\'e} et~al.(2011)Andr{\'e}, Vercauteren, Buchner, Wallace, and
  Ayache]{andre2011smart}
B.~Andr{\'e}, T.~Vercauteren, A.~M. Buchner, M.~B. Wallace, and N.~Ayache.
\newblock A smart atlas for endomicroscopy using automated video retrieval.
\newblock \emph{Medical image analysis}, 15\penalty0 (4):\penalty0 460--476,
  2011.

\bibitem[Arjovsky and Bottou(2017)]{arjovsky2017towards}
M.~Arjovsky and L.~Bottou.
\newblock Towards principled methods for training generative adversarial
  networks.
\newblock \emph{stat}, 1050:\penalty0 17, 2017.

\bibitem[Ayasso et~al.(2012)Ayasso, Rodet, and Abergel]{ayasso2012variational}
H.~Ayasso, T.~Rodet, and A.~Abergel.
\newblock A variational bayesian approach for unsupervised super-resolution
  using mixture models of point and smooth sources applied to astrophysical
  map-making.
\newblock \emph{Inverse Problems}, 28\penalty0 (12):\penalty0 125005, 2012.

\bibitem[Bao et~al.(2017)Bao, Chen, Wen, Li, and Hua]{bao2017cvae}
J.~Bao, D.~Chen, F.~Wen, H.~Li, and G.~Hua.
\newblock Cvae-gan: Fine-grained image generation through asymmetric training.
\newblock In \emph{Proceedings of the IEEE Conference on Computer Vision and
  Pattern Recognition}, pages 2745--2754, 2017.

\bibitem[Bulat et~al.(2018)Bulat, Yang, and Tzimiropoulos]{bulat2018learn}
A.~Bulat, J.~Yang, and G.~Tzimiropoulos.
\newblock To learn image super-resolution, use a gan to learn how to do image
  degradation first.
\newblock \emph{arXiv preprint arXiv:1807.11458}, 2018.

\bibitem[Goodfellow et~al.(2014)Goodfellow, Pouget-Abadie, Mirza, Xu,
  Warde-Farley, Ozair, Courville, and Bengio]{goodfellow2014generative}
I.~Goodfellow, J.~Pouget-Abadie, M.~Mirza, B.~Xu, D.~Warde-Farley, S.~Ozair,
  A.~Courville, and Y.~Bengio.
\newblock Generative adversarial nets.
\newblock In \emph{Advances in neural information processing systems}, pages
  2672--2680, 2014.

\bibitem[Greenspan(2008)]{greenspan2008super}
H.~Greenspan.
\newblock Super-resolution in medical imaging.
\newblock \emph{The Computer Journal}, 52\penalty0 (1):\penalty0 43--63, 2008.

\bibitem[Johnson et~al.(2016)Johnson, Alahi, and
  Fei-Fei]{johnson2016perceptual}
J.~Johnson, A.~Alahi, and L.~Fei-Fei.
\newblock Perceptual losses for real-time style transfer and super-resolution.
\newblock In \emph{European Conference on Computer Vision}, pages 694--711.
  Springer, 2016.

\bibitem[Kennedy et~al.(2006)Kennedy, Israel, Frenkel, Bar-Shalom, and
  Azhari]{kennedy2006super}
J.~A. Kennedy, O.~Israel, A.~Frenkel, R.~Bar-Shalom, and H.~Azhari.
\newblock Super-resolution in pet imaging.
\newblock \emph{IEEE transactions on medical imaging}, 25\penalty0
  (2):\penalty0 137--147, 2006.

\bibitem[Kennedy et~al.(2007)Kennedy, Israel, Frenkel, Bar-Shalom, and
  Azhari]{kennedy2007improved}
J.~A. Kennedy, O.~Israel, A.~Frenkel, R.~Bar-Shalom, and H.~Azhari.
\newblock Improved image fusion in pet/ct using hybrid image reconstruction and
  super-resolution.
\newblock \emph{International journal of biomedical imaging}, 2007, 2007.

\bibitem[Ledig et~al.(2017)Ledig, Theis, Huszar, Caballero, Cunningham, Acosta,
  Aitken, Tejani, Totz, Wang, et~al.]{ledig2016photo}
C.~Ledig, L.~Theis, F.~Huszar, J.~Caballero, A.~Cunningham, A.~Acosta,
  A.~Aitken, A.~Tejani, J.~Totz, Z.~Wang, et~al.
\newblock Photo-realistic single image super-resolution using a generative
  adversarial network.
\newblock In \emph{Proceedings of the IEEE Conference on Computer Vision and
  Pattern Recognition}, pages 4681--4690, 2017.

\bibitem[Matkovic et~al.(2005)Matkovic, Neumann, Neumann, Psik, and
  Purgathofer]{matkovic2005global}
K.~Matkovic, L.~Neumann, A.~Neumann, T.~Psik, and W.~Purgathofer.
\newblock {Global Contrast Factor} - a {New Approach to Image Contrast}.
\newblock \emph{Computational Aesthetics}, 2005:\penalty0 159--168, 2005.

\bibitem[Mescheder et~al.(2017)Mescheder, Nowozin, and
  Geiger]{mescheder2017adversarial}
L.~Mescheder, S.~Nowozin, and A.~Geiger.
\newblock Adversarial variational bayes: Unifying variational autoencoders and
  generative adversarial networks.
\newblock In \emph{International Conference on Machine Learning}, pages
  2391--2400, 2017.

\bibitem[Neumann et~al.(2010)Neumann, Kiesslich, Wallace, and
  Neurath]{neumann2010confocal}
H.~Neumann, R.~Kiesslich, M.~B. Wallace, and M.~F. Neurath.
\newblock Confocal laser endomicroscopy: technical advances and clinical
  applications.
\newblock \emph{Gastroenterology}, 139\penalty0 (2):\penalty0 388--392, 2010.

\bibitem[Nguyen et~al.(2015)Nguyen, Lee, Parekh, Samarasena, Bechtold, and
  Chang]{nguyen2015current}
D.~L. Nguyen, J.~G. Lee, N.~K. Parekh, J.~Samarasena, M.~L. Bechtold, and
  K.~Chang.
\newblock The current and future role of endomicroscopy in the management of
  inflammatory bowel disease.
\newblock \emph{Annals of gastroenterology: quarterly publication of the
  Hellenic Society of Gastroenterology}, 28\penalty0 (3):\penalty0 331, 2015.

\bibitem[Plenge et~al.(2012)Plenge, Poot, Bernsen, Kotek, Houston, Wielopolski,
  van~der Weerd, Niessen, and Meijering]{plenge2012super}
E.~Plenge, D.~H. Poot, M.~Bernsen, G.~Kotek, G.~Houston, P.~Wielopolski,
  L.~van~der Weerd, W.~J. Niessen, and E.~Meijering.
\newblock Super-resolution reconstruction in mri: better images faster?
\newblock In \emph{Medical Imaging 2012: Image Processing}, volume 8314, page
  83143V. International Society for Optics and Photonics, 2012.

\bibitem[Rav{\`\i} et~al.(2017)Rav{\`\i}, Fabelo, Callic, and
  Yang]{ravi2017manifold}
D.~Rav{\`\i}, H.~Fabelo, G.~M. Callic, and G.-Z. Yang.
\newblock Manifold embedding and semantic segmentation for intraoperative
  guidance with hyperspectral brain imaging.
\newblock \emph{IEEE transactions on medical imaging}, 36\penalty0
  (9):\penalty0 1845--1857, 2017.

\bibitem[Rav{\`\i} et~al.(2018)Rav{\`\i}, Szczotka, Shakir, Pereira, and
  Vercauteren]{ravi}
D.~Rav{\`\i}, A.~B. Szczotka, D.~I. Shakir, S.~P. Pereira, and T.~Vercauteren.
\newblock Effective deep learning training for single-image super-resolution in
  endomicroscopy exploiting video-registration-based reconstruction.
\newblock \emph{IJCARS (in press), March}, 2018.

\bibitem[Sano et~al.(2017)Sano, Mori, Goto, Hirano, and
  Funahashi]{sano2017super}
Y.~Sano, T.~Mori, T.~Goto, S.~Hirano, and K.~Funahashi.
\newblock Super-resolution method and its application to medical image
  processing.
\newblock In \emph{Consumer Electronics (GCCE), 2017 IEEE 6th Global Conference
  on}, pages 1--2. IEEE, 2017.

\bibitem[Shocher et~al.(2017)Shocher, Cohen, and Irani]{shocher2017zero}
A.~Shocher, N.~Cohen, and M.~Irani.
\newblock " zero-shot" super-resolution using deep internal learning.
\newblock \emph{arXiv preprint arXiv:1712.06087}, 2017.

\bibitem[Torre et~al.(2016)Torre, Siegel, Ward, and Jemal]{torre2016global}
L.~A. Torre, R.~L. Siegel, E.~M. Ward, and A.~Jemal.
\newblock Global cancer incidence and mortality rates and trends--an update.
\newblock \emph{Cancer Epidemiology and Prevention Biomarkers}, 25\penalty0
  (1):\penalty0 16--27, 2016.

\bibitem[Villena et~al.(2013)Villena, Vega, Babacan, Molina, and
  Katsaggelos]{villena2013bayesian}
S.~Villena, M.~Vega, S.~D. Babacan, R.~Molina, and A.~K. Katsaggelos.
\newblock Bayesian combination of sparse and non-sparse priors in image super
  resolution.
\newblock \emph{Digital Signal Processing}, 23\penalty0 (2):\penalty0 530--541,
  2013.

\bibitem[Wang et~al.(2004)Wang, Bovik, Sheikh, and Simoncelli]{wang2004image}
Z.~Wang, A.~C. Bovik, H.~R. Sheikh, and E.~P. Simoncelli.
\newblock Image quality assessment: from error visibility to structural
  similarity.
\newblock \emph{IEEE transactions on image processing}, 13\penalty0
  (4):\penalty0 600--612, 2004.

\bibitem[Xiao et~al.(2010)Xiao, Hays, Ehinger, Oliva, and
  Torralba]{xiao2010sun}
J.~Xiao, J.~Hays, K.~A. Ehinger, A.~Oliva, and A.~Torralba.
\newblock Sun database: Large-scale scene recognition from abbey to zoo.
\newblock In \emph{Computer vision and pattern recognition, 2010 IEEE
  conference on}, pages 3485--3492. IEEE, 2010.

\bibitem[Zhu et~al.(2017)Zhu, Park, Isola, and Efros]{zhu2017unpaired}
J.-Y. Zhu, T.~Park, P.~Isola, and A.~A. Efros.
\newblock Unpaired image-to-image translation using cycle-consistent
  adversarial networks.
\newblock In \emph{Proceedings of the IEEE Conference on Computer Vision and
  Pattern Recognition}, pages 2223--2232, 2017.

\end{thebibliography}


\bibliographystyle{abbrvnat}


\end{document}